\documentclass{article}
\usepackage{arxiv}
\usepackage{wrapfig}
\usepackage{subcaption}
\usepackage{makecell}
\usepackage{amsmath}
\usepackage{graphicx}
\usepackage{graphbox}

\title{player2vec: A Language Modeling Approach to Understand Player Behavior in Games}

\author{
    Tianze Wang\thanks{Both authors contributed equally to this research.}\\
	KTH Royal Institute of Technology\\
	Sweden\\
	\texttt{tianzew@kth.seu}\\
    \And
    Maryam Honari-Jahromi\footnotemark[1]\\
	King\\
	Sweden\\
	\texttt{maryam.honari@king.com} \\
	\AND
    Styliani Katsarou\\
	King\\
	Sweden\\
    \texttt{stella.katsarou@king.com}
	\And
    Olga Mikheeva\\
	King and KTH Royal Institute of Technology\\
	Sweden\\
    \texttt{olga.mikheeva@king.com}
    \And
    Theodoros Panagiotakopoulos\\
	King\\
	Sweden\\
    \texttt{theodoros.panagiotakopoulos@king.com} \\
    \AND
    Sahar Asadi\thanks{Joint senior authorship.}\\
	King\\
	Sweden\\
    \texttt{sahar.asadi@king.com}
    \And
    Oleg Smirnov\footnotemark[2]\\
	King\\
	Sweden\\
    \texttt{oleg.smirnov@king.com}
}

\begin{document}
\date{}
\maketitle

\begin{abstract}
Methods for learning latent user representations from historical behavior logs have gained traction for recommendation tasks in e-commerce, content streaming, and other settings. However, this area still remains relatively underexplored in video and mobile gaming contexts. In this work, we present a novel method for overcoming this limitation by extending a long-range Transformer model from the natural language processing domain to player behavior data. We discuss specifics of behavior tracking in games and propose preprocessing and tokenization approaches by viewing in-game events in an analogous way to words in sentences, thus enabling learning player representations in a self-supervised manner in the absence of ground-truth annotations.
We experimentally demonstrate the efficacy of the proposed approach in fitting the distribution of behavior events by evaluating intrinsic language modeling metrics. Furthermore, we qualitatively analyze the emerging structure of the learned embedding space and show its value for generating insights into behavior patterns to inform downstream applications.
\end{abstract}

\section{Introduction}
\label{sec:introduction}
User modeling and user understanding are important ingredients to deliver high-quality customer experience in many online platforms, ranging from music and video streaming to retail and e-commerce~\cite{zhou2018atrank, hariri2012context}, to health informatics~\cite{feely2023modelling}. Example downstream applications include understanding user journey and conversion funnel~\cite{aly2012web}, delivering personalized content recommendations~\cite{tian2020survey}, classifying intents~\cite{lu2020knowledge}, and predicting lifetime value and churn~\cite{bauer2021improved,ahn2020survey} among others. Traditional approaches that rely on explicit annotations collected from user surveys are expensive to scale to modern platforms with millions of users. As a result, researchers and practitioners in this field often turn to learning from implicit signals~\cite{sun2019bert4rec,luo2023mcm}, such as behavior tracking data that can be sourced from online interaction logs and is available in abundant amounts. Due to the lack of ground-truth labels, training is framed in a self-supervised learning paradigm, where a model is pretrained on a carefully constructed \emph{pretext} objective. In this scenario, pseudo labels are derived from the inherent structure of data, allowing the algorithm to learn the underlying distribution, thereby serving as advantageous initializations for a wide range of downstream tasks. Denoising autoencoder is a successful example of self-supervised objectives in the natural language processing (NLP) domain, where a model is trained to predict randomly omitted tokens from a sequence, which serve as classification pseudo labels~\cite{Lewis2019BARTDS}.

Despite being largely successful in other settings, large-scale modeling of user behavior remains relatively untapped in the gaming domain, where previous studies were mainly focused on exploratory analysis~\cite{drachen2013game}, psychology-informed supervised learning~\cite{pedersen2009modeling,yannakakis2013player}, as well as clustering and other unsupervised methods~\cite{drachen2012guns,siqueira2017data}. In this work, we aim to bridge this gap and investigate the applicability of self-supervised learning from tracking data with the purpose of modeling \emph{player} behavior in casual mobile games. Inspired by recent advances in the natural language domain, we adopt Longformer~\cite{beltagy2020longformer}, a variant of Transformer~\cite{vaswani2017attention} architecture tailored to long context lengths, along with the masked language modeling (MLM) objective~\cite{kenton2019bert} that has proven successful in various sequence modeling methods.

Similar to conventional language modeling approaches, we exploit pretraining a model on a large unlabeled corpus to learn latent representations that are applicable to downstream tasks. We hypothesize common patterns in how individual players interact with game content and mechanics can add a new dimension to player understanding.

In this paper, we propose a novel approach to understanding player behavior in the mobile gaming domain by extending long-range Transformer models from the NLP domain to create a context-rich representation of the player from sessionized raw player behavior logs in a self-supervised manner. In Section~\ref{sec:dataset}, we describe the player behavior data collected from a large mobile gaming platform. We outline the challenges in modeling raw data and propose a preprocessing pipeline in Section~\ref{sec:preprocessing}. Then, we introduce the model architecture and training procedure in Section~\ref{sec:modeling}. Quantitative and qualitative experimental results are presented in Section~\ref{sec:experiments}. Finally, we conclude and discuss potential directions for future work in Section~\ref{sec:conclusion}.

\section{Related Work}
\label{sec:related_work}
Modeling sequential interactions between users and items was explored in prior works for recommendation tasks. Earlier, learning from temporal data for collaborative filtering was investigated under a simplifying Markovian assumption~\cite{zimdars2001using}, which was later extended to a Markov decision process~\cite{shani2005mdp}. Predicting future behavioral trajectories using contextual and sequential information was addressed with an autoregressive Long Short-Term Memory model~\cite{wu2017recurrent}. Another research introduced a coupled Recurrent Neural Network architecture to jointly model the user/item interactions trajectory in the embedding space~\cite{kumar2019predicting}. Furthermore, it has been shown that explicitly modeling different types of user behavior, such as repeated consumption and various actions, can improve the performance in downstream metrics~\cite{anderson2014dynamics, ren2019repeatnet}.

Leveraging language models for embedding sequential data in recommendation settings was pioneered in the context of learning music track representations~\cite{mehrotra2018towards} with Word2Vec objective~\cite{mikolov2013efficient} from user-generated playlists. Later, this approach was extended to modeling sequences of listening sessions with a Recurrent Neural Network~\cite{hansen2020contextual}.

More recently, a self-attention sequential model was introduced~\cite{sun2019bert4rec}, where the authors argued that the Transformer framework finds the optimal trade-off between simpler Markov chain models, which typically excel in high-sparsity settings, and neural network methods that can capture more complex scenarios but require dense training data. In follow-up work, a multi-task customer model for personalization~\cite{luo2023mcm} outperformed the previous state-of-the-art BERT4Rec model~\cite{sun2019bert4rec} by leveraging a novel data augmentation and task-aware readout module.

However, the applications of language models for user modeling remain understudied for gaming-specific tasks. To the best of our knowledge, we propose the first approach for learning representations of mobile game players based on pretraining a Transfomer architecture in a self-supervised manner.

\section{Methodology}
\label{sec:methodology}
Player behavior data, while being superficially similar to tracking data in other domains, possesses several notable differences that make it challenging to model with conventional approaches. First, since video games are designed to be a dynamic and engaging environment, in-game user interactions happen at a much higher frequency than in Web browsing and other applications. Thereafter, it leads to large amounts of potentially redundant behavior events that require careful data preprocessing and modeling long-range dependencies for capturing informative patterns. 

Second, player behavior data often contain a specific type of noise that is not present and, hence, not well-studied in other scenarios. Specifically, a fraction of events in a given time window may have wrong ordering or do not contain any ordering information. This problem typically manifests due to real-world engineering and operations constraints, for example, users switching between online and offline playing modes. This noise can deteriorate model performance during training and inference.

\subsection{Dataset}
\label{sec:dataset}
When a player interacts with a mobile game application, their behavior generates a sequence of time-ordered events, which are recorded locally on a user's device and sent to the central game server. Example events include starting the application, starting a new game round, purchasing in-game items, and displaying pop-ups and notifications from the application side, among others. Supported behavior events are grouped in a vocabulary of $12$ semantic classes, where each class is uniquely identified by its name and has an associated event schema that consists of a collection of mandatory and optional continuous and categorical features. 

\begin{figure}[h]
    \centering
    \includegraphics[width=0.75\textwidth]{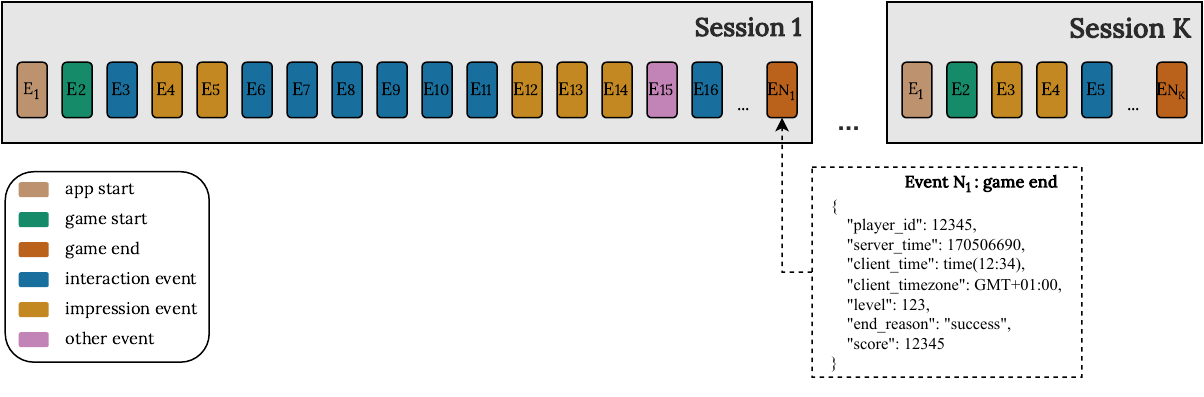}
    \caption{An illustration of the categorization of events into sessions. The final event in session 1, marked as a game-end event, is expanded to show details about its associated fields and values.}
    \label{fig:sessions}
\end{figure}

\begin{figure}[ht]
    \centering
    \begin{subfigure}[t]{0.30\textwidth}
        \includegraphics[trim=0.2cm 0.2cm 0.7cm 0.2cm, clip,align=c,height=4cm]{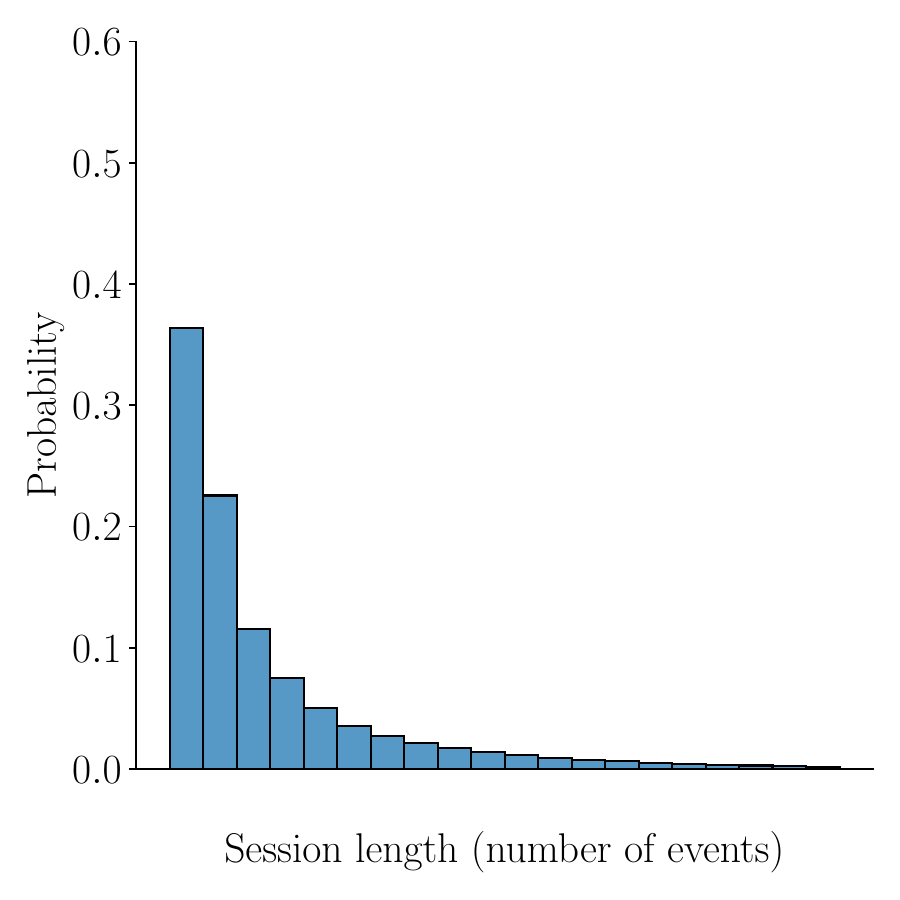}
        \caption{}
        \label{fig:dataset_session_length}
    \end{subfigure}
    ~
    \begin{subfigure}[t]{0.30\textwidth}
        \includegraphics[trim=0.2cm 0.2cm 0.7cm 0.2cm, clip,align=c,height=4cm]{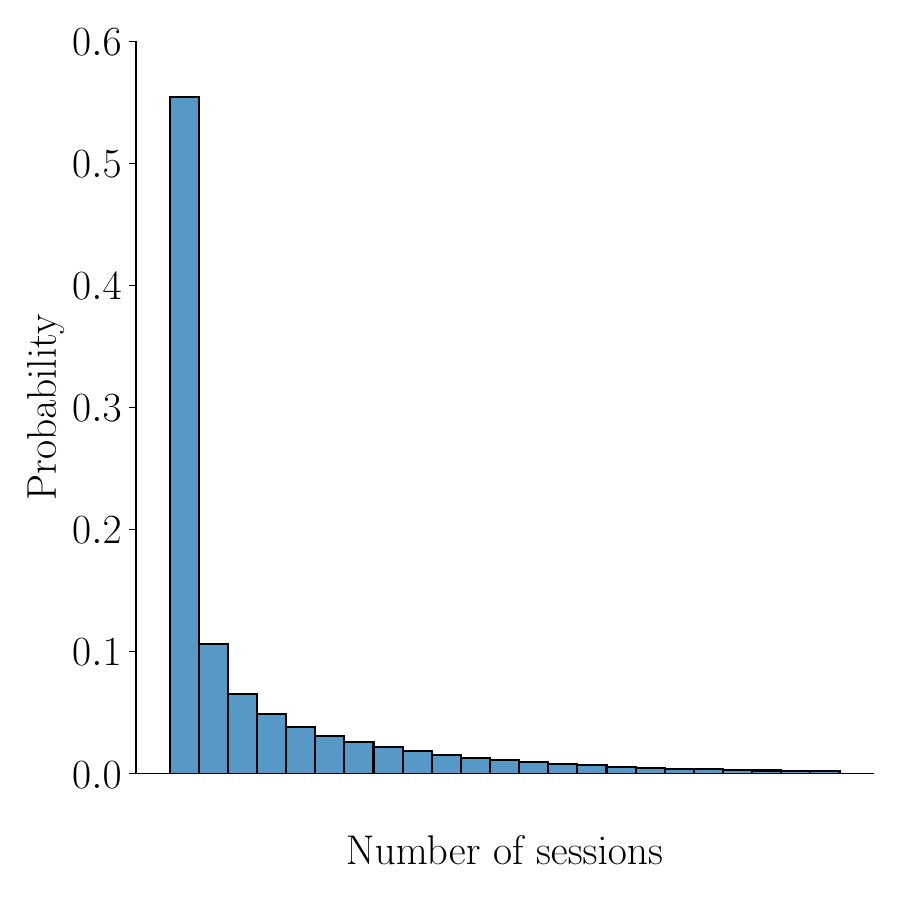}
        \caption{}
        \label{fig:dataset_activity_dist}
    \end{subfigure}
    ~
    \begin{subfigure}[t]{0.30\textwidth}
        \includegraphics[trim=0.5cm 0.2cm 0.2cm 0.2cm, clip,align=c,height=4cm]{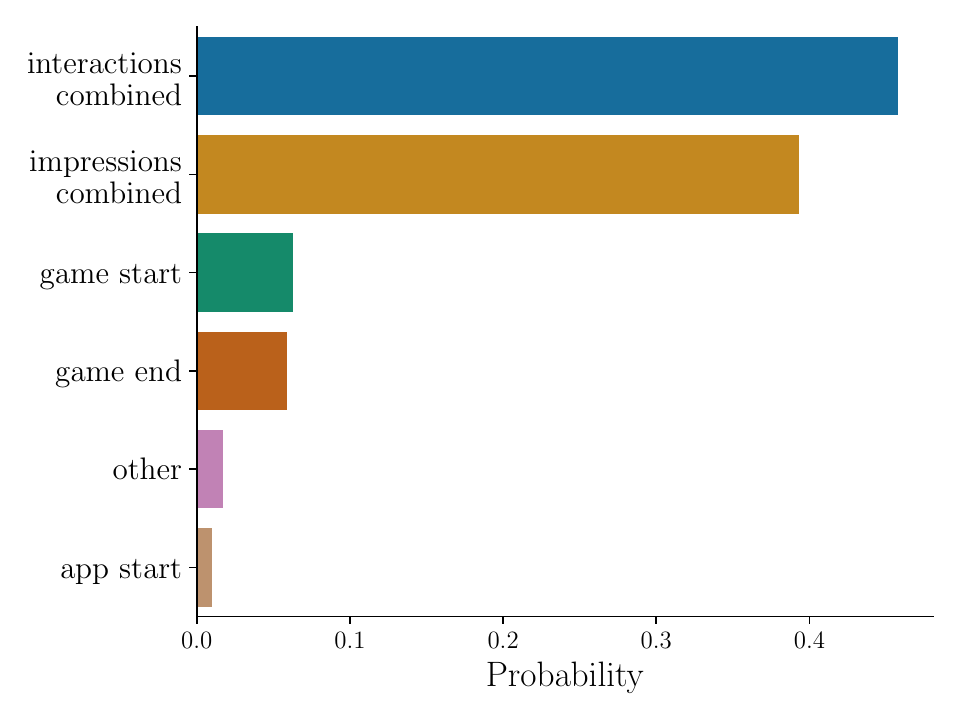}
        \caption{}
        \label{fig:dataset_event_distribution}
    \end{subfigure}
    \caption{(a) Histogram of session lengths in the dataset. (b) Distribution of player activity over a 15-day period. (c) Event distribution, where events belonging to similar semantic classes are grouped together. Plots in (a) and (b) show data up to the 99\textsuperscript{th} percentile.}

\end{figure}

Player behavior events are grouped into sessions based on player activity, as illustrated in Figure~\ref{fig:sessions}. An instance of a \texttt{game end} event is depicted in the figure, with proprietary information removed. Given the product domain knowledge and prior analysis, we consider a session as ended if a player is inactive for $15$ minutes. The distribution of session lengths in player data is shown in Figure~\ref{fig:dataset_session_length}. Figure~\ref{fig:dataset_activity_dist} depicts the distribution of the number of sessions of active players over a period of 15 days. We observe that both session lengths and player activities approximately follow geometric distribution, which is expected for this kind of data.

As illustrated in Figure~\ref{fig:dataset_event_distribution}, some types of events are much more prevalent in the data. This imbalance in the distribution can create a bias problem and skew the model's predictions unless accounted for in the preprocessing stages. Notably, although the data is sequential, the event structure is non-trivial to map into tokens, hence allowing many possible design choices that we discuss in the next section.

In this work, we use a dataset of player behavior sessions collected from a large mobile game provider over 15 days with $10{,}000$ players uniformly sampled from our user base. The resulting database consists of $125{,}000$ sessions, where $67\%$ is allocated for the training split and the rest $33\%$ for the validation split.

\subsection{Preprocessing}
\label{sec:preprocessing}
Player behavior events collected from game applications are contained in structured JSON logs. In order to leverage approaches from natural language modeling, we design a pipeline to transform raw events into rich yet compact textual sequences that a model can adequately consume.

\begin{figure}[t]
    \centering
    \includegraphics[width=\textwidth]{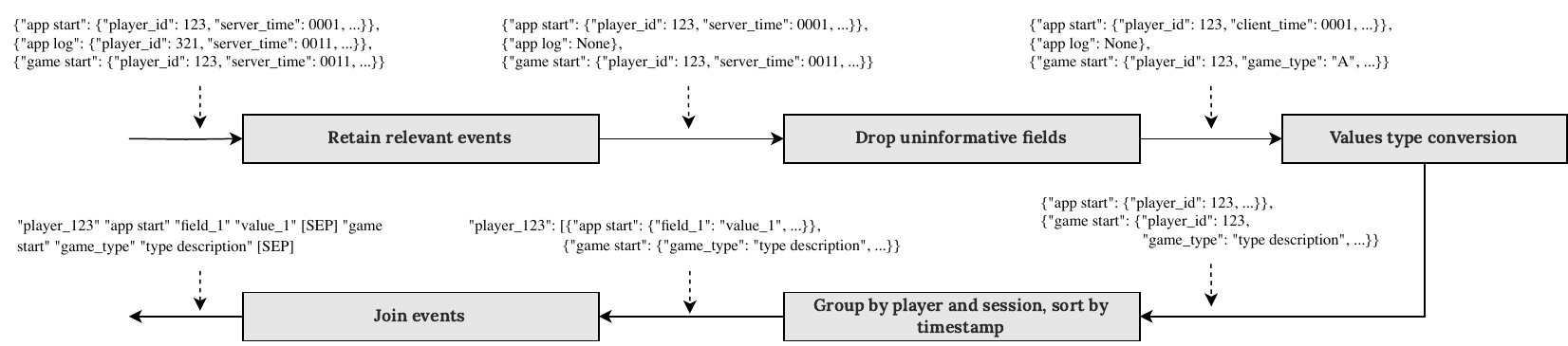}
    \caption{Data preprocessing pipeline. Raw event logs are passed through filtering, type conversion, grouping, and joining stages to produce textual data.}
    \label{fig:preprocessing_diagram}
\end{figure}

The preprocessing pipeline, as shown in Figure~\ref{fig:preprocessing_diagram}, initially eliminates superfluous events and event fields. We build upon product knowledge to filter out data that is not informative for the player's behavior, such as device-specific logs, where many fields are mere artifacts of the application internals. At this stage, the number of event fields is reduced on average by over 90\%. Following, raw numerical values and identifiers are converted to text, assessing a curated dictionary that maps atomic identities to descriptive terms. For instance, all identifiers corresponding to social activities are condensed to ``social'' token. In the next stage, events are grouped by users and sessions and ordered by their timestamps to preserve the narrative flow of interactions. Finally, individual behavior sessions are concatenated to form a textual representation of a game player.

We observe a fraction of the events in the raw data do not contain correct chronological ordering. This issue arises due to the logging platform limitations in situations when players are interacting with the application offline, which prevents online communications with the game server. We theorize that this type of position noise can hinder the performance of language model-based methods, which are inductively biased toward learning from consecutive tokens. However, to enrich our analysis and maintain a realistic scenario that allows for ordering noise, we chose a dataset that underwent minimal preprocessing, retaining a higher proportion of mixed events compared to a version reduced to less than $1\%$ noise by means of additional preprocessing steps.

We make use of a word-level tokenizer that splits a space-separated input string into individual tokens and then maps tokens to unique identifiers. Word-level tokenization is motivated by a relatively small vocabulary size of behavior data (${\sim}13{,}500$ tokens) compared to the vocabularies of natural languages. However, the resulting tokenized sequences are significantly longer than those that are generally used in NLP tasks, such as sentiment analysis and question answering.

\subsection{Modeling}
\label{sec:modeling}
Modeling long sequences poses a serious challenge to Transformer-based approaches due to self-attention operation, which scales quadratically with the input length in memory and computational complexity. As a result, sequences longer than $512$ tokens that are commonly found in player behavior data remain out of reach for conventional BERT~\cite{kenton2019bert} architecture and its derivatives.
The problem is further exacerbated by modeling distant dependencies in extended playing history that involves concatenating multiple sessions. To address this limitation, we adopt Longformer~\cite{beltagy2020longformer}, a model that is specifically designed for processing long documents. Longformer relies on a combination of dilated sliding window attention to capture local context and global attention on a few pre-selected input locations. This formulation scales linearly with the input size, enabling the processing of sequences up to $4{,}096$ tokens in a single pass, which is sufficient in most behavior modeling scenarios. Moreover, Longformer's sparse attention pattern excels in settings where a portion of tokens in the immediate local context are expected to be redundant, which is the case for high-frequency gaming behavior data.

We validate the proposed approach using several model variants with different capacities using the hyperparameters listed in Table~\ref{tbl:hyperparams}. We experiment with the baseline Longformer configuration, denoted as \emph{player2vec-large}, as well as two smaller model size variants with reduced numbers of internal layers and self-attention heads.

\begin{table}[ht]
\begin{center}
\begin{tabular}{lrrrrr} 
 Model variant & Hidden layers & \#Heads & Hidden dimensions & Block size & 

\#Param ($\sim  $)\\ 
 \hline
 \emph{player2vec-small} & 2 & 2 & 128 & 1024& 2M \\ 
 \emph{player2vec-medium} & 6 & 6 & 384 & 2048 & 20M \\ 
 \emph{player2vec-large} & 12 & 12 & 768 & 4096 & 121M \\ 
\end{tabular}
\caption{Model hyperparameters per model size variant.}
\label{tbl:hyperparams}
\end{center}
\end{table}

\section{Experiments}
\label{sec:experiments}
To ensure a fair comparison in all experiments, we train each network from randomly initialized weights for 100 epochs with a batch size of $4$ and gradient accumulation over $4$ steps, which results in an effective batch size of $16$ ($2^{16}$ tokens). We also apply weight decay of $10^{-2}$ on all parameters for regularization. The models were optimized with MLM objective by Adam algorithm~\cite{kingma2014adam} with a fixed learning rate $2 \times 10^{-5}$ throughout training. We use HuggingFace Transformers~\cite{wolf2019huggingface} library and PyTorch framework~\cite{paszke2019pytorch} for model implementation. All networks were trained with the half-precision (FP16) format on a single NVIDIA A100 GPU accelerator, where the largest model takes $\approx50h$ to train. 

\subsection{Intrinsic performance}
\label{sec:performance}
First, we evaluate the goodness of fit of the proposed approach with intrinsic MLM metrics. We report cross-entropy and multi-class classification accuracy of predicting masked tokens computed on the validation split for tested model architectures in Table~\ref{tbl:metrics}. We also report the perplexity score following the existing methodology for evaluating MLM pretraining performance~\cite{liu2019roberta}. As expected, we observe that a larger capacity model is able to fit the behavior sessions more accurately while not entering the overfitting regime.

\begin{table}[ht]
\begin{center}
\begin{tabular}{lrrrrr} 
 Model variant & Block size & Accuracy~$\uparrow$ & Perplexity~$\downarrow$ & Cross-entropy~$\downarrow$ \\ [0.5ex] 
\hline
 \emph{player2vec-small} & 1024 &  $ 0.698 \pm 0.067 $ & $ 3.272 \pm 0.714 $ &  $ 1.161 \pm 0.222$ \\ 
 \emph{player2vec-medium} & 2048 & $ 0.934 \pm 0.015 $ & $ 1.287 \pm 0.093 $ & $ 0.250 \pm 0.069$  \\ 
 \emph{player2vec-large} & 4096 &  $ \mathbf{0.958} \pm \mathbf{0.007} $ & $ \mathbf{1.161} \pm \mathbf{0.046} $ & $\mathbf{0.149}\pm \mathbf{0.040}$ \\ 

\end{tabular}

\caption{Masked language modeling intrinsic metrics mean values and standard deviations computed over 5 training runs.}
\label{tbl:metrics}
\end{center}
\end{table}

\subsection{Cluster analysis}
\label{sec:clusters}
\begin{figure}[t]

    \centering\
     \begin{subfigure}[t]{0.41\textwidth} 
        \includegraphics
        [trim=0.2cm 1cm 0cm 0.2cm, clip, align=l, width=\textwidth]
        {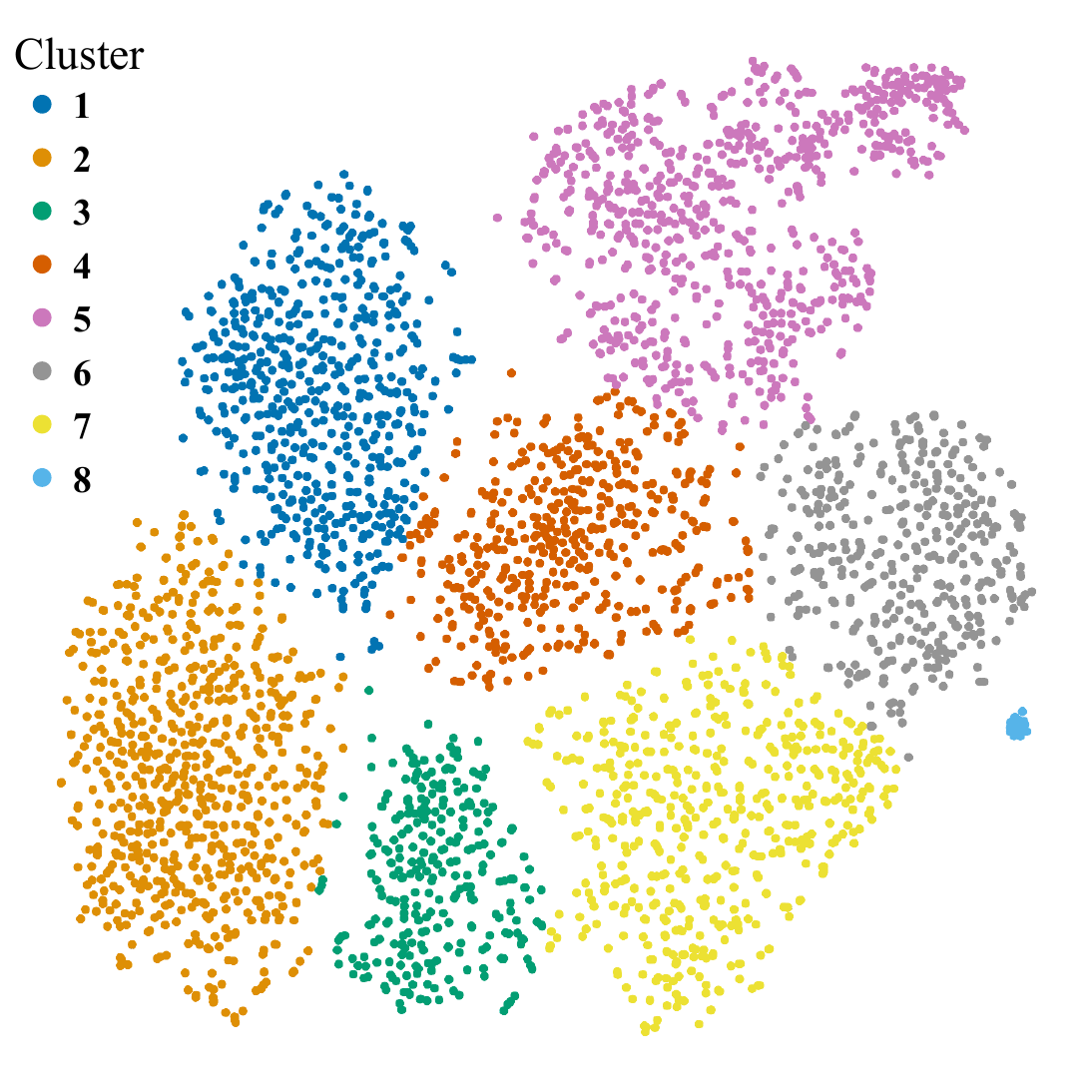}
        \caption{} \label{fig:clustering-tsne}
    \end{subfigure}
    \hfill
    ~
     \begin{subfigure}[t]{0.46\textwidth} 
        \includegraphics[trim=-1cm 0.1cm 0cm 0cm, clip, align=r, width=\textwidth]{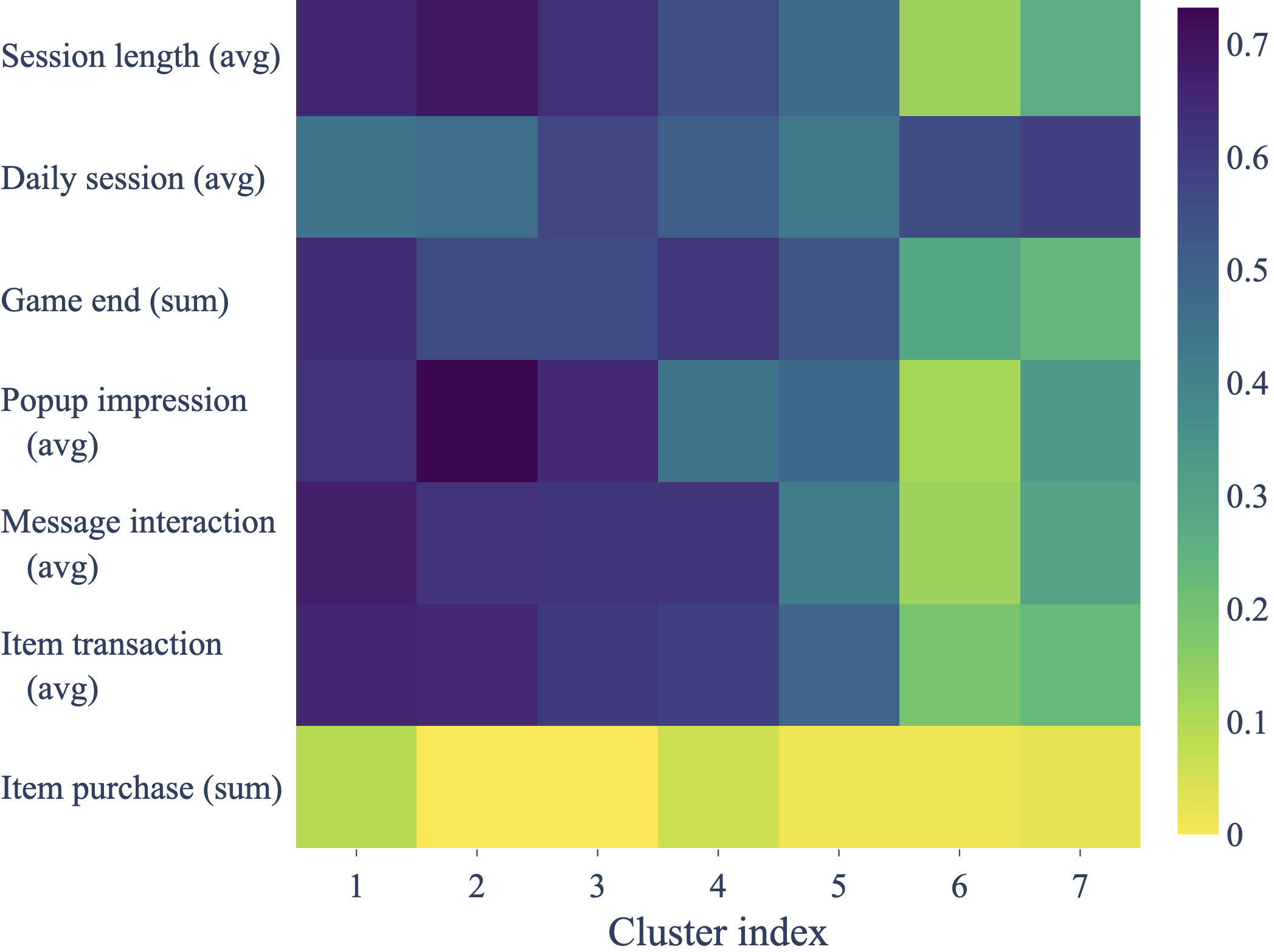}
        \caption{} \label{fig:clustering-fingerprint}
    \end{subfigure}
    ~
    \caption{(a) t-SNE of latent embedding space obtained from pre-trained player2vec-large with subsequent GMM clustering. (b) Histogram of the quantized player events in identified clusters. We exclude cluster 8 due to the small cluster size and no gameplay.
    }
    \label{fig:tsne_plot}
\end{figure}

Next, we perform qualitative analysis by visualizing a t-SNE~\cite{van2008visualizing} plot of the learned latent space to identify clusters that are representative of player behavior. To this end, we extract embeddings of input token sequences from the converged \emph{player2vec} large variant that showed the strongest performance with respect to MLM accuracy. We use $4096$ by $768$-dimensional representations produced by the last Transformer layer, which is further aggregated with max pooling operation over sequence length to compute an embedding vector for a player input sequence. We additionally project obtained session embeddings onto the first 50 principal components with linear PCA method to suppress noise and speed up computation. The obtained projections are then mapped to 2D space via t-SNE and clustered using Gaussian Mixture Model~\cite{gmmreynolds2009} (8 components). The resulting t-SNE plot is depicted in Figure~\ref{fig:clustering-tsne}.

Analyzing the average player behavior in the well-separated t-SNE clusters in Figure~\ref{fig:clustering-fingerprint}, we rediscover known segments in the players' population from the product point of view:

\begin{enumerate}
\item \textbf{competitive devoted}: a highly skilled player who plays less frequently on a daily basis but for long sessions, occasionally purchases items, and collects utilities.
\item \textbf{casual devoted} -- a resourceful player who plays long sessions but not frequently, engages in-game quests, enjoys collecting utilities and rewards, but prefers the free gameplay experience.
\item \textbf{persistent devoted} -- a resourceful player who plays frequent and long sessions while enjoying the free gameplay experience.
\item \textbf{lean-in casual economy aware} --  a skilled player who plays less often but longer sessions, collects and occasionally purchases items.
\item \textbf{lean-in casual} -- a skilled player who plays less often but has longer sessions.
\item \textbf{persistent casual} -- a not very skillful player who plays short and frequent sessions while being less engaged in social and economic aspects of the game.
\item \textbf{persistent collector} -- a player with frequent short sessions, collecting utilities to pass the levels.
\end{enumerate}

We conclude that the main components of variation in the embedding space meaningfully correspond to the high-level player behavior segmentation. This observation underscores the potential utility of the embeddings for downstream tasks. Furthermore, this type of analysis cannot capture the entire essence of a player due to some attributes not being clearly presented by a set of features. We will extend the features to reflect features highlighted in player insights studies. Regardless, this method of examining player segments offers an interesting perspective to game designers and calls for a deeper understanding of users' behavior and motivation.

\section{Ethical Considerations}
\label{sec:ethics}
Computational modeling of players in games has raised numerous concerns in research and industrial communities~\cite{mikkelsen2017ethical}. In this work, we use non-personally identifiable tracking data recorded from in-game interactions to create vectorized representations of player behaviors. Our goal is to leverage such representations to support personalized and enhanced player experiences. We identify potential ethical risks that can arise from i) bias in the input dataset, e.g., underrepresenting less frequent players data, or ii) misapplication of well-validated models to a different data distribution, e.g., using a model trained on expert players for new players, known as Type III errors~\cite{mikkelsen2017ethical}. To tackle those challenges, we leverage tooling for data validation and automated model analysis, which are available in production-ready machine learning frameworks~\cite{46484}. Additionally, we capture underrepresented or misrepresented player behavior by employing qualitative evaluation methods such as embedding space visualization. We address distribution shifts by periodically retraining the model with recent data, where the retraining cadence is determined empirically based on performance and distribution drift. We envision applying model explainability and uncertainty estimation methods on the downstream recommendation system to understand better the model's robustness, biases, and other ethical considerations. 

\section{Conclusion and Future Work}
\label{sec:conclusion}
This paper introduces a novel user behavior modeling approach inspired by language modeling principles. To generate player embeddings, we utilize tracking events that form sessions, mirroring the way how word tokens in NLP compose sentences and paragraphs. We showcase a method for modeling player behavior data self-supervised by pretraining a Transformer-based architecture with long context length on a dataset of tracking events in the gaming domain. We experimentally demonstrated the efficacy of the proposed pipeline by evaluating intrinsic MLM metrics. Moreover, we qualitatively analyzed the emerging structure of the learned embedding space extracted from a pre-trained model. We showed the existence of semantic structures, such as clusters of users based on their play styles and in-game spending behavior, that further support the viability of the proposed method for player representation learning. We discovered previously unknown user subpopulations that serve as valuable insights into downstream product applications. 

For the next steps, we plan to extend the training procedure to single- and multitask fine-tuning with labeled datasets to benchmark the model performance against fully-supervised baselines. Furthermore, in future work, we will focus on quantifying and mitigating the effect of position-based noise on representation learning tasks that are commonly present in mobile game settings.

\section*{Acknowledgements}
The authors would like to
thank Gabriela Zarzar Gandler and Bj\"{o}rn Brinne,
who helped with the conceptualization of the project in the early phases,
and Labinot Polisi, Martin Lundholm, and Dionysis Varelas, who assisted with
acquiring player behavior data in a live production environment.

\bibliographystyle{acm}
\bibliography{main}
\end{document}